%% file: main.tex
\documentclass[conference]{IEEEtran}
\IEEEoverridecommandlockouts
\usepackage{cite}
\usepackage{amsmath,amssymb,amsfonts}
\usepackage{algorithmic}
\usepackage{graphicx}
\usepackage{textcomp}
\usepackage{xcolor}
\usepackage{textcomp}
\usepackage{subcaption}
\usepackage[T1]{fontenc}
\usepackage[numbers]{natbib}
\def\BibTeX{{\rm B\kern-.05em{\sc i\kern-.025em b}\kern-.08em
    T\kern-.1667em\lower.7ex\hbox{E}\kern-.125emX}}
\begin{document}
\input{symbols}
\title{Deciphering Human Mobility: Inferring Semantics of Trajectories with Large Language Models\\
\thanks{\IEEEauthorrefmark{2}These authors contributed equally to this work.

\IEEEauthorrefmark{1}Corresponding author}
}

\author{
\IEEEauthorblockN{Yuxiao Luo\IEEEauthorrefmark{2}, Zhongcai Cao\IEEEauthorrefmark{2}, Xin Jin, Kang Liu\IEEEauthorrefmark{1}, Ling Yin\IEEEauthorrefmark{1}}
\IEEEauthorblockA{\textit{Shenzhen Institute of Advanced Technology} \\
\textit{Chinese Academy of Sciences}\\
Shenzhen, China \\
\{yx.luo1, zc.cao, xin.jin1, kang.liu, yinling\}@siat.ac.cn}
}

\maketitle

\begin{abstract}
Understanding human mobility patterns is essential for various applications, from urban planning to public safety. The individual trajectory such as mobile phone location data, while rich in spatio-temporal information, often lacks semantic detail, limiting its utility for in-depth mobility analysis. Existing methods can infer basic routine activity sequences from this data, lacking depth in understanding complex human behaviors and users' characteristics. Additionally, they struggle with the dependency on hard-to-obtain auxiliary datasets like travel surveys. To address these limitations, this paper defines trajectory semantic inference through three key dimensions: user occupation category, activity sequence, and trajectory description, and proposes the Trajectory Semantic Inference with Large Language Models (TSI-LLM) framework to leverage LLMs infer trajectory semantics comprehensively and deeply. We adopt spatio-temporal attributes enhanced data formatting (STFormat) and design a context-inclusive prompt, enabling LLMs to more effectively interpret and infer the semantics of trajectory data. Experimental validation on real-world trajectory datasets demonstrates the efficacy of TSI-LLM in deciphering complex human mobility patterns. This study explores the potential of LLMs in enhancing the semantic analysis of trajectory data, paving the way for more sophisticated and accessible human mobility research.
\end{abstract}

\begin{IEEEkeywords}
Human mobility analysis, Large language models, Trajectory semantic inference.
\end{IEEEkeywords}
\input{section/1_introduction}
\input{section/2_related_work}
\input{section/4_method}

\input{section/5_experiment}
\input{section/6_results}

\input{section/7_conclusion}
\input{section/8_acknowledgement}
\bibliographystyle{IEEEtranN}
\bibliography{IEEEfull}

\end{document}

%% file: symbols.tex
\newcommand{\inputseq}{X}
\newcommand{\poivector}{\mathbf{H}}
\newcommand{\tfidf}{w}
\newcommand{\poitypenum}{N}

%% file: section/1_introduction.tex
\section{Introduction}
A better understanding of human mobility is pivotal across diverse applications, including the improvement of urban planning~\cite{haraguchi2022human}, optimization of transportation systems~\cite{yuan2011t}, control of infectious diseases~\cite{liu2020enhancing}, enhancement of public safety measures~\cite{zhu2021agent}, and so on. With the advent of advanced information and communication technologies (ICTs), individual trajectory data such as mobile phone location tracking data has become the most used data type for human mobility analysis. Nevertheless, mobile phone location data records typically do not contain specific semantic details (e.g., the activity type such as working or leisure). This lack of information hinders the accurate interpretation and utilization of the collected data for various human mobility studies and applications.

To uncover the semantics embedded within trajectories, researchers have suggested advanced models that utilize auxiliary datasets~\cite{widhalm2015discovering} to deduce them as activity sequences~\cite{bowman2001activity}. For example,~\citet{diao2016inferring} constructed a multinomial logit regression model using travel survey data for forecasting semantic labels.~\cite{yin2021mining} proposes a flexible white box method to infer activity purposes based on spatial and temporal features with the probability model. However, the above models suffer from some limitations. Firstly, most of these methods need auxiliary datasets (e.g., travel surveys) for improving the preformance, which can be challenging to gather and may not be accessible to all users. Secondly,  inferring semantics results in merely identifying a sequence of regular activities (e.g., Home, Work, and School) without delving into a deeper comprehension of the rich semantics behind human spatio-temporal behaviors and user characteristics, such as occupations and accociated activities.
\begin{figure}[t]
  \centering
  \includegraphics[width=0.45\textwidth]{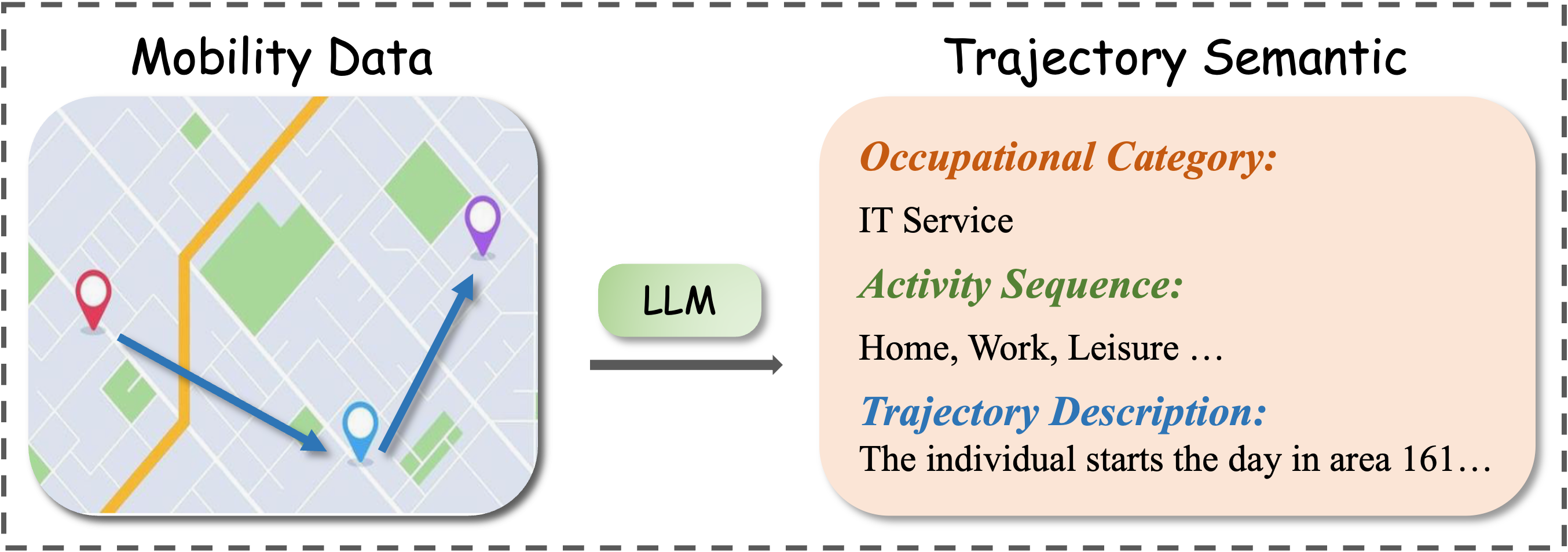}
  \caption{LLM-based trajectory semantic inference.}
  \label{fig:intro}
\end{figure}

To address the above issues, this study initially defines trajectory semantic inference from three dimensions and introduces a novel inference framework based on Large Language Models (LLMs) as shown in Figure \ref{fig:intro}. For comprehensive deciphering of human mobility from individual trajectories, we suggest that the trajectory semantic inferences should involve three components. These components include the user occupational category, which significantly influences mobility behavior~\cite{wang2019extended}, the activity sequence, serving as a structured representation of activity semantics, and the trajectory description, which encompasses the context of activity purpose and essential spatio-temporal information. Based on the recently launched LLMs~\cite{achiam2023gpt,team2023gemini,touvron2023llama} with extensive knowledge and reasoning capacity, we proposed a novelty \textbf{T}rajectory \textbf{S}emantic \textbf{I}nference framework with \textbf{L}arge \textbf{L}anguage \textbf{M}odels (\textbf{TSI-LLM}) to infer the semantics of individual trajectory data and decipher human mobility from the above three dimensions without difficult-to-access ancillary datasets like the travel survey. To enhance the understanding of individual trajectories by LLMs, we employ the spatio-temporal attributes enhanced data formatting (STFormat) with the group-based sampling strategy to reprogram the trajectory data as sptial-temporal trajectory chain with abundant semantic information. Moreover, we carefully design context-inclusive prompts to aid in inferring trajectory semantics within TSI-LLM. The main contributions are listed as follows:
\begin{itemize} 
    \item We define trajectory semantic inference using three dimensions: occupational category, activity sequence, and trajectory description.
    \item A novelty Trajectory Semantic Inference with Large Language Models (TSI-LLM) is introduced, incorporating spatio-temporal attributes enhanced data formatting (STFormat) and trajectory semantics inference prompt for comprehensive analysis and inference of individual trajectory semantics. 
    \item We conduct experiments using real trajectory data to demonstrate the effectiveness of TSI-LLM.
\end{itemize}
\begin{figure*}[t]
  \centering
  \includegraphics[width=0.88\textwidth]{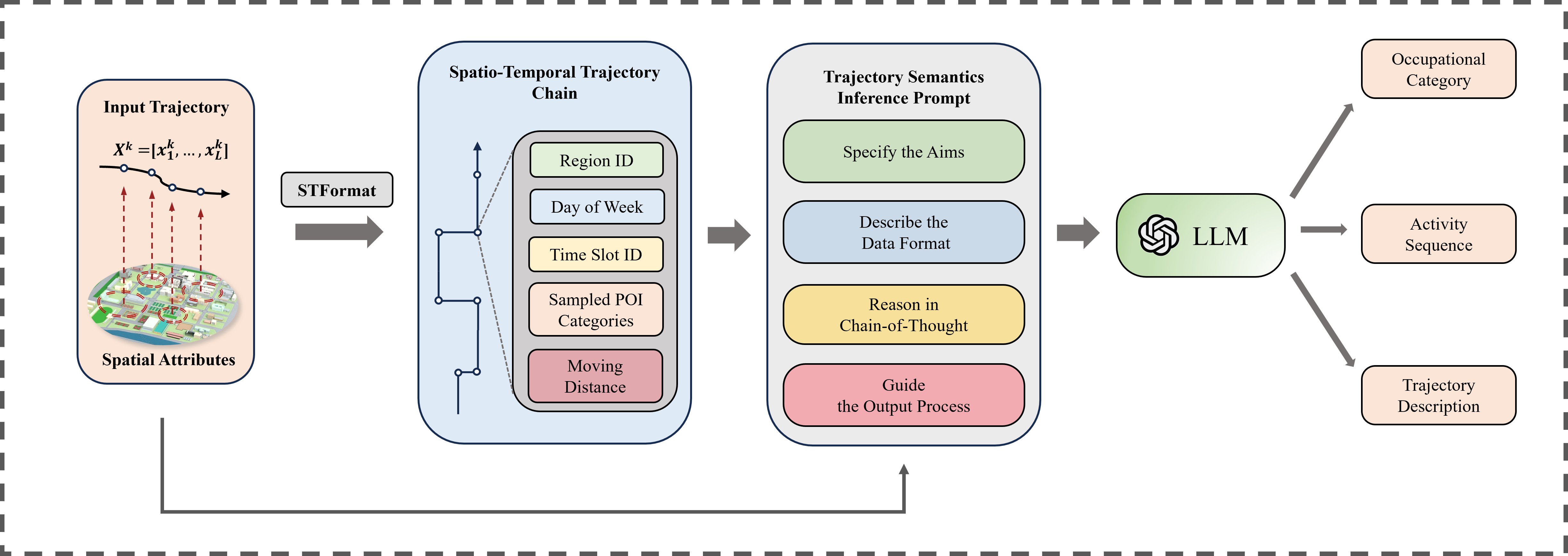}
  \caption{Framework of TSI-LLM: The raw trajectory is transformed into the spatio-temporal trajectory chain through spatio-temporal attributes enhanced data formatting (STFormat). The trajectory semantics inference prompt guides LLMs to process the spatio-temporal trajectory chain and trajectory data. Following the instructions, LLMs finally output three inference results: occupational category, activity sequence, and trajectory description.}
  \label{fig:TSI_LLM}
\end{figure*}

%% file: section/2_related_work.tex
\section{Related Work}
\subsection{Inferring Trajectory Semantics through Activity Sequence}

 Understanding human mobility is crucial for a wide range of applications, e.g. advancing urban planning~\cite{haraguchi2022human}, optimizing transportation systems~\cite{yuan2011t}, and managing infectious diseases~\cite{liu2020enhancing}. The trajectory semantics of individuals like activity information offer an important understanding of people's daily movements and behaviors in different environments~\cite{bowman2001activity}. The activity sequence is a structured representation of trajectory semantics that typically denotes a sequence of daily activities~\cite{kitamura1988evaluation}. However, obtaining or inferring the activity sequence information is challenging~\cite{yin2021mining}. Conventional methods for gathering activity sequence data typically rely on travel surveys, which can be resource-intensive and lack instantaneity and continuity~\cite{mcdonald2008critical}. Advancements in mobile communication technologies now allow for the tracking of an individual's daily movement locations, and activity semantics embedded in the trajectories can be inferred. For instance,~\citet{gong2016inferring} introduced a Bayesian rule-based approach to deduce trip purposes from taxi trajectory information, taking into account spatial and temporal limitations.~\citet{yin2017generative} proposed an enhanced hidden Markov network (HMN) model that takes into account the transition patterns among various activities in order to deduce activity sequence for daily commuters.~\citet{yin2021mining} introduced a versatile white box approach to extract sequences of human activities through spatial and temporal characteristics of daily activities. Although these models can identify regular activities, they often lack complicated human spatiotemporal behaviors and user characteristics. Furthermore, they rely on auxiliary datasets which are typically only accessible to specific transportation agencies.

\subsection{Human Mobility Analysis Using Large Language Models}
Large Language Models (LLMs), such as GPT-4~\cite{achiam2023gpt}, Gemini~\cite{team2023gemini}, and Llama~\cite{touvron2023llama} have shown promise in capturing complex linguistic patterns and generating coherent text across various domains. 
Recently, there has been research exploring the potential of utilizing LLMs in studies related to human mobility. For example, LLM-Mob leveraged the language understanding and reasoning capabilities of LLMs incorporating context-inclusive prompts to predict the next location of the trajectory~\cite{wang2023would}. UrbanGPT integrated a spatiotemporal dependency encoder with the instruction-tuning paradigm, enhancing its ability to predict unseen urban traffic flow accurately~\cite{li2024urbangpt}. MobiGeaR utilizes a divide-and-coordinate mechanism that leverages the synergistic effect of LLM reasoning and the mechanistic gravity model to generate human mobility behavior~\cite{shao2024beyond}. However, research has yet to explore the use of LLMs for inferring the semantics of individual trajectories in human mobility analysis. We are making the first attempt to use LLMs to connect trajectory data with its underlying semantics, inferring the users' activity purposes and related characteristics.

%% file: section/4_method.tex
\section{Methodology}
In this section, we introduce the framework of our proposed Trajectory Semantic Inference with Large Language Models (TSI-LLM) in detail. As illustrated in Figure 1, our approach to enhance LLMs' comprehension of individual trajectories involves two main phases: spatio-temporal attributes enhanced data formatting (STFormat) and trajectory semantics inference prompt design. In the first phase, we utilize both spatial and temporal information to construct a spatio-temporal trajectory chain of the raw trajectory data. This includes Points of Interest (POI) in the regions visited by users, the distances traveled, and the related timestamp. In the second phase, we design prompts specifically to aid LLMs in understanding the formatted data and in inferring the trajectory semantics, which includes occupational category, activity sequence, and trajectory description.

\subsection{Preliminaries}
An individual trajectory $\inputseq^k$ is defined as a sequence $[x_1^k,\dots,x_L^k ]$. And $x_l^k$ indicates the ID of a specific spatial unit or region where individual $k$ spent the majority of his/her time during the equal interval time slot $l$ (e.g., 9:00-9:59). The semantic inference of the trajectory problem we introduced aims to derive three types of analytical outcomes: occupational category, activity sequence, and trajectory description from the raw trajectory $\inputseq^k$. 

\subsection{Spatio-temporal Attributes Enhanced Data Formatting}
The raw trajectory data $\inputseq^k$ is a sequence of region ID, however utilizing the raw data is not sufficient for meaningful trajectory analysis. Therefore, we enrich this data by incorporating additional spatio-temporal attributes. Daily human activities are intricately connected to spatial characteristics such as land use types and building types in their vicinity~\cite{yang2019revealing,yin2021mining}, which represent spatial information within the region. To enhance the capability of LLMs to understand the spatial features and link them to activity purposes, we adopt a classical method TF-IDF~\cite{yuan2012discovering} to formulate a POI feature vector $\poivector_r$ of each region.  $\poivector_r$ is donated by $[\tfidf_1,\tfidf_2,\dots,\tfidf_M ]$, where $\tfidf_i$ is the TF-IDF value of the i-th POI category. It can be formalized as:
\begin{equation}
    \tfidf_i=\frac{n_i}{N_r}\times\log\frac R{\left\|\{q|\text{the i-th POI category }\in q\}\right\|},
\end{equation}
where $n_i$ represents the count of POIs within the i-th category, and $N_r$ denotes the overall number of POIs in that specific region. $R$ stands for the total count of regions in the city, and $q$ indicates the number of regions containing the ith POI category. Then sampling the primary POI category through $\poivector_r$ in the region visited by the user can reflect the potential activity.

However, in a highly developed modern city, the categories of POIs are diverse and extensive, spanning from commercial establishments such as shopping malls and restaurants to cultural venues like museums and theaters. A single type of POI may not fully capture the diversity and richness of user preferences and behaviors in urban spaces. Meanwhile, because of the significant variation in quantity distribution, some important POI categories with low weight are rarely included in the samples. In this case, we propose a group-based sampling strategy to tackle this problem. Initially, we group all POI categories based on their urban function and associated action purpose into $\poitypenum$ predetermined basic function groups. There are five groups of POI categories following~\citet{yin2021mining}, involving Home, Work, School, Leisure, and Other. Then $\poivector_{r,n}$ represents the weight vector of the POI category for the n-th group within the r-th region. The process of group-based sampling can be formalized as:
\begin{equation}
\begin{split}
    {\Tilde{\poivector}_{r,n}}&=Softmax(\poivector_{r,n}),\\
    S_{r,n}&=Multinomial({\Tilde{\poivector}_{r,n}},K),\\
    S_r&=cat(S_{r,1},\dots,S_{r,N}),
\end{split}
\end{equation}
where the Softmax function transforms the weight vector $\poivector_{r,n}$  into a probability distribution $\Tilde{\poivector}_{r,n}$, the Multinomial function then samples K elements, representing the selected categories, and $cat$ represents a concatenation operation. $S_r$ consisting of $N\times K$ elements is the final sampled POI categories within r-th region. Using this method, we aim to ensure a more balanced representation of various basic functional groups in our sampling process, allowing for a more comprehensive understanding of user activity in urban spaces. 

The user's traveling distance between stayed regions in two consecutive time slots represents another vital spatial attribute to enrich the trajectory data, which is implicit in interregional spatial information. We add this distance as auxiliary information. Meanwhile, the temporal factor also significantly affects the user's activity purpose~\cite{zhang2018understanding}. The daily routines of residents are frequently influenced by societal norms, leading to regular daily habits. In this case, we utilize the stay time interval index (0-23) and work day of the week (1-5) as the timestamp for the trajectory sequence, to help LLMs use temporal information to infer the intended activities at that moment. Finally, after STFormat, the raw trajectory data will be transformed into a spatio-temporal trajectory chain as Figure \ref{fig:TSI_LLM}. 

\begin{figure}[!tb]
  \centering
  \includegraphics[width=0.34\textwidth]{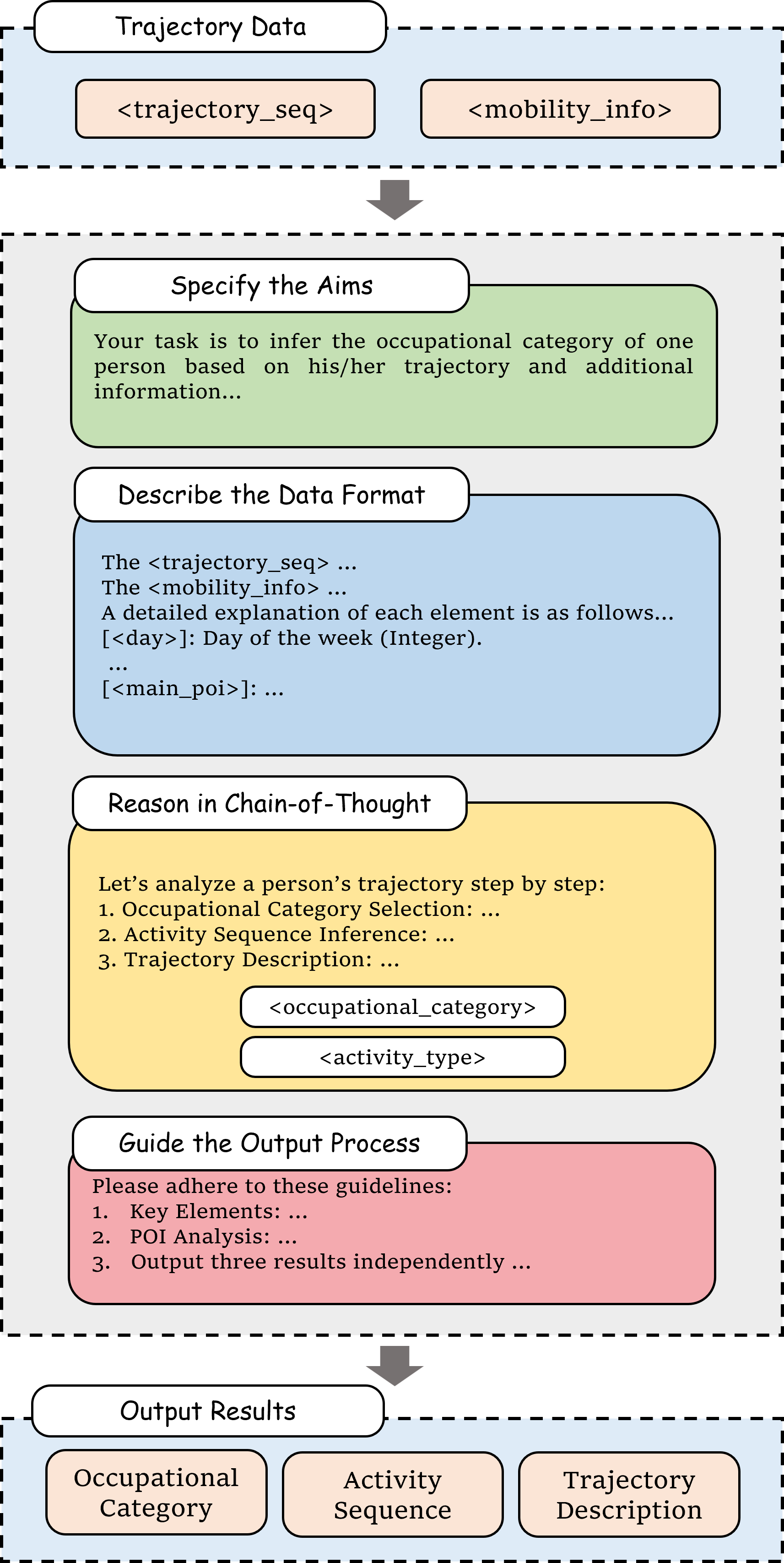}
  \caption{The content of trajectory semantics inference prompt involves specifying the aims, describing the data, reasoning in a chain-of-thought, and guiding the output.}
  \label{fig:prompt}
\end{figure}

\subsection{Trajectory Semantics Inference Prompt Design}
The prompt design significantly impacts LLMs performance, influencing both knowledge retrieval accuracy and computational efficiency~\cite{linzbach2023decoding}. Expanding on established prompting techniques like Chain-of-Thought (CoT)~\cite{wei2022chain}, we develop context-inclusive prompts that integrate pertinent contextual details to improve LLMs' comprehension of individual trajectories. As depicted in Figure \ref{fig:prompt}, it is comprised of various components, each annotated according to its function.

\paragraph{Specify the Aims}
\label{sepcify_aim}
In our proposed prompt, we begin by outlining the tasks and specifying our input trajectory data and mobility information using two unique placeholders, <trjactory\_seq> and <mobility\_info>. Here, <trjactory\_seq> is the trajectory sequence formed by reign IDs the individual stayed and <mobility\_info> is the spatio-temporal trajectory chain. To comprehensively infer the trajectory semantics as thoroughly as possible, we set three sub-tasks: Occupational Category Selection, Activity Sequence Inference, and Trajectory Description. Initially, the instruction asks the model to classify the individual's occupation by analyzing their movement patterns and correlating these with specific occupational categories. Next, we expect to infer the sequence of activities undertaken by the individual, deducing their daily routine and behavioral patterns from their spatial and temporal movement data. Finally, the model needs a detailed narrative of the individual's trajectory, encapsulating the entirety of their movements and elucidating the purpose and context of each segment. Due to the extensive length of mobility information, we employ the <mobility\_info> placeholder to represent the entirety of this data in the prompt text, ensuring the coherence of the instructions is maintained. Furthermore, it is essential to provide a description of the data sources from which cite are drawn to facilitate the retrieval of relevant prior knowledge for LLMs.

\paragraph{Describe the Data}
We present a comprehensive description of the two types of input data, trajectory sequence and mobility information, aiming to facilitate the comprehension of this information for LLMs. The trajectory sequence represents an individual's movement path in a consistent format of region IDs arranged chronologically. Furthermore, we have elaborated on the structure of mobility information, highlighting the importance of each component and data type.

\paragraph{Reason in Chain-of-Thought}
To ensure the model infers reasonable semantics of trajectory, we follow the CoT strategy to decompose the task as multi-step problems. Initially, we request the model to identify the user's occupation from predefined <occupational\_category>, as this is critical for analyzing their likely movement patterns and destinations~\cite{wang2019extended}. Once the user's occupation is determined, the model then narrows down the potential range of activities associated with that occupation. Then, based on the potential occupation category and mobility information, the model needs to infer the activity type of this person at each region from predefine <activity\_type> \cite{yin2021mining}. We predefine five activity types including Home, Work, School, Leisure, and Other. Subsequently, the model details the trajectory by utilizing the identified occupational category and activity sequence, analyzing it to elucidate the behavior patterns at specific times and locations. This step-by-step approach allows the model to logically analyze the trajectory and output information across three levels: Occupational Category, Activity Sequence, and Trajectory Description.

\paragraph{Guide the Output Process}
There are some guidelines provided to standardize the model output. Specifically, the model should emphasize key time points, active areas, and moving distances in its analysis. Additionally, the description must clarify the purpose of activities in each stay region based on the main POI type. It is important to note that the model is expected to infer the three most likely scenarios as results, because a single trajectory could potentially depict multiple outcomes.

%% file: section/5_experiment.tex
\section{Experiment}
In this section, we evaluate the inference performance and reasonableness of TSI-LLM on a real individual trajectory dataset.
\subsection{Datasets and Preprocessing}
We chose Shenzhen City, China as the study area to validate the effectiveness of our proposed method. The raw trajectory data was sourced from the Smart Steps\footnote{http://www.smartsteps.com}, a big data company under China Unicom, and each original raw trajectory is composed of a sequence of “stay” locations with start and end time. We randomly select 100 samples from different anonymous individuals from 1 November to 7 November 2021 without their personally identifiable information. To achieve an appropriate spatiotemporal resolution, the raw locations defined by latitude and longitude will be converted into certain spatial units, specifically neighborhoods (i.e., "Shequ" in Chinese). Additionally, trajectories will be segmented into time slots based on duration of stay. This mapping approach is designed to lessen the computational burden while preserving adequate analytical accuracy. The POI dataset\footnote{www.amap.com} includes approximately 1,500,000 records with locations and covers 122 categories.
\subsection{Implementation Details}
The specific LLM employed in the experiment is GPT-4 with open APIs\footnote{https://platform.openai.com/docs/models/gpt-4-and-gpt-4-turbo}, which is one of the most advanced LLMs. We set the temperature to 0.1 for less randomness in the output. The input trajectory length $L$ is set as 24 (1 day). The number $K$ of POI categories group-based sampling is set as 3 for each group, which means there are 15 ($5 \times 3$) sampled POI categories for each region in a trajectory.

%% file: section/6_results.tex
\begin{figure}[t]
  \centering
  \begin{subfigure}[b]{0.38\textwidth}
    \includegraphics[width=\textwidth]{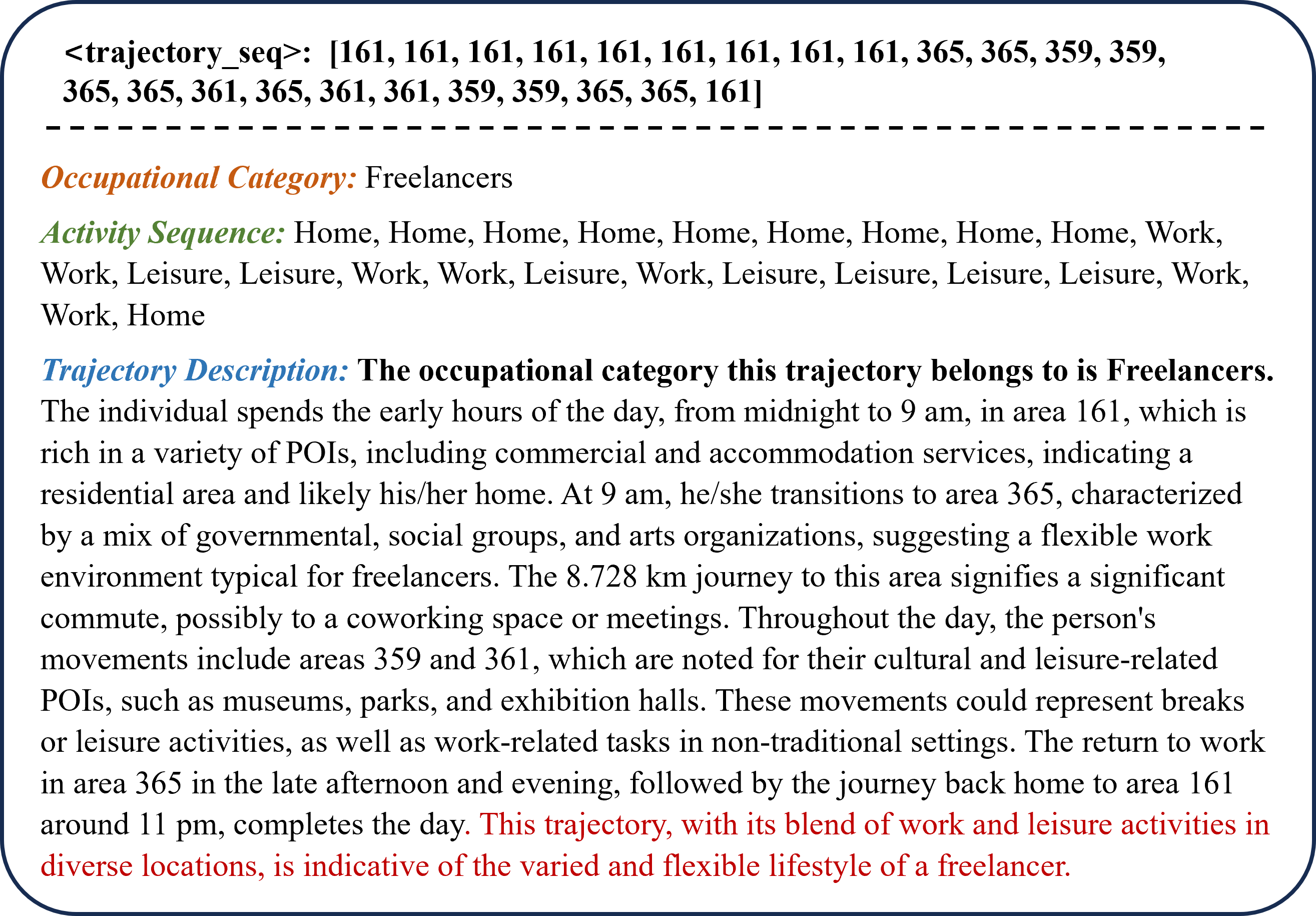}
    \caption{Result A}
  \end{subfigure}
  \begin{subfigure}[b]{0.38\textwidth}
    \includegraphics[width=\textwidth]{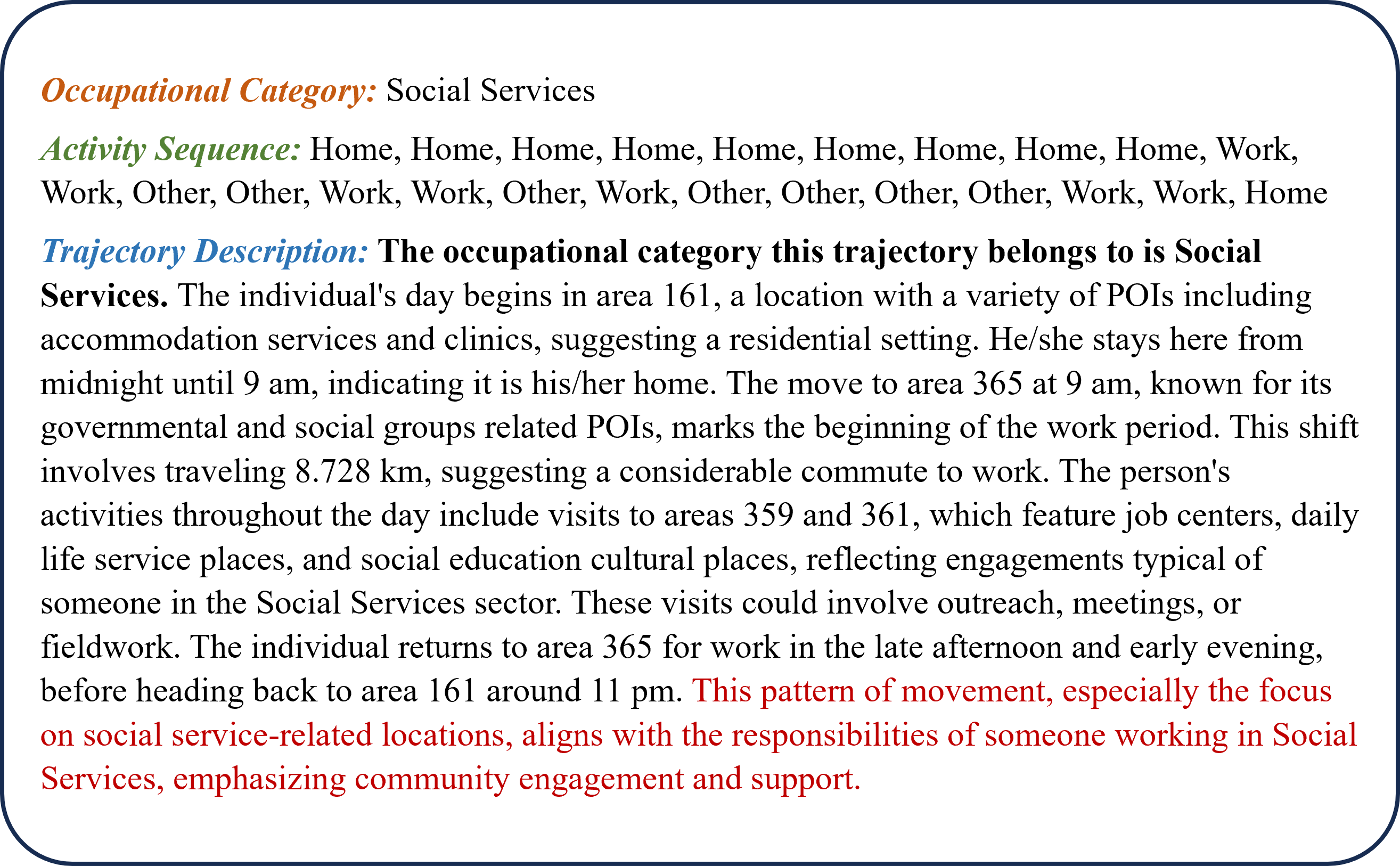}
    \caption{Result B}
  \end{subfigure}
  \begin{subfigure}[b]{0.38\textwidth}
    \includegraphics[width=\textwidth]{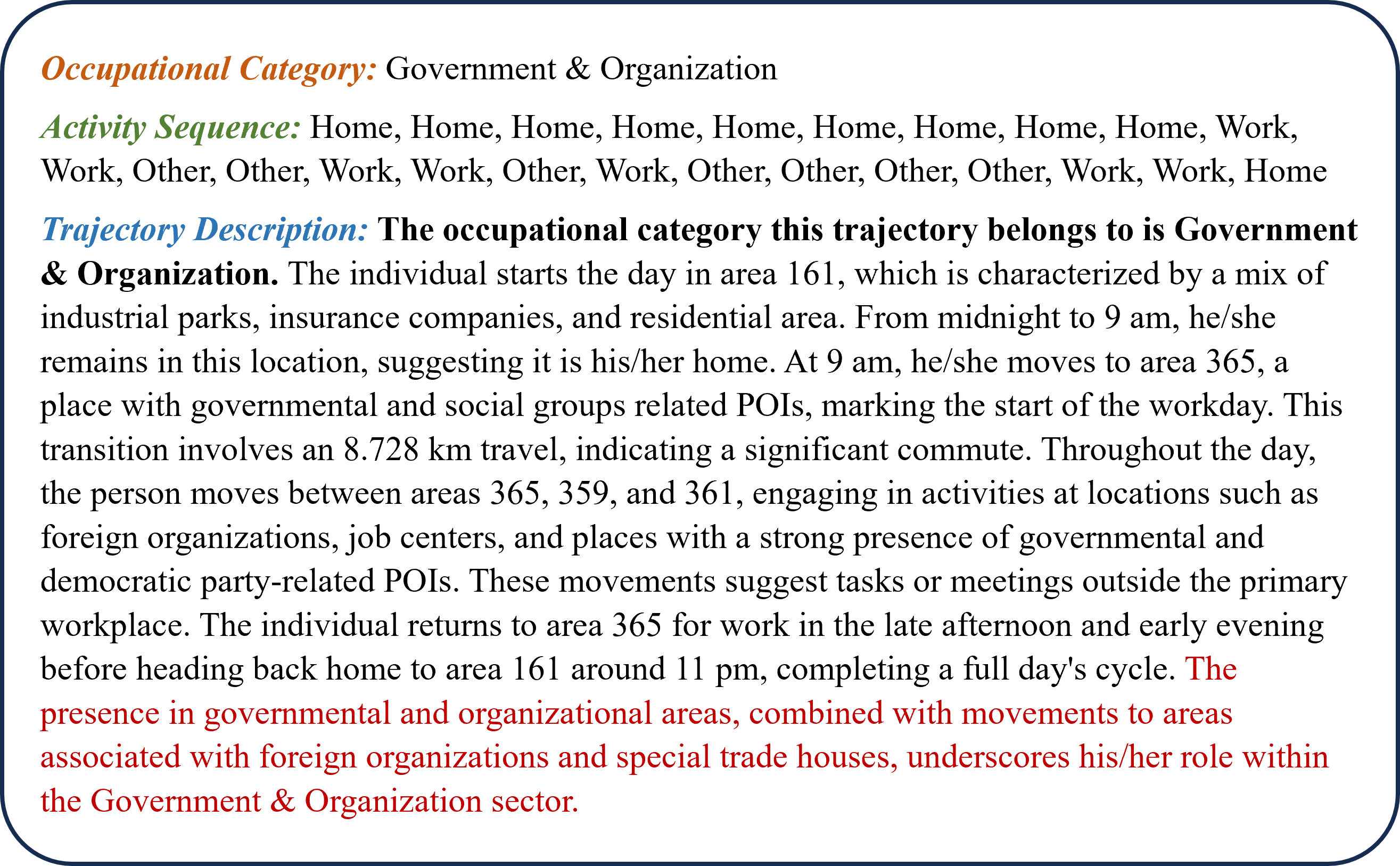}
    \caption{Result C}
  \end{subfigure}
  \caption{The example of the TSI-LLM for semantic inference of the individual trajectory. Each result is comprised of three distinct outputs: Occupational Category, Activity Sequence and Trajectory Description.}
  \label{fig:results}
\end{figure}
\section{Results}

As for the results, we check the 100 samples carefully. Due to the page limitation, we choose one sample as an example to demonstrate the rationality of inference result from TSI-LLM. 
In this example, the trajectory sequence is [161, 161, 161, 161, 161, 161, 161, 161, 161, 365, 365, 359, 359, 365, 365, 361, 365, 361, 361, 359, 359, 365, 365, 161], we input the corresponding <trajectory\_seq> and <mobility\_info> combined with prompt into LLM, and obtained three different results.

As shown in Figure \ref{fig:results}, TSI-LLM gives three different categories of individuals based on the given trajectory and conducts a reasonable analysis of the activities of the individuals. TSI-LLM categorizes the individual into three distinct occupational categories based on the given trajectory and conducts a reasonable analysis of the activities of the individual. For each inference result, TSI-LLM first identifies the occupational category of the individual. Then, combining the occupational category and mobility information, it infers the activities carried out by the individual at each moment to form an activity sequence. Finally, it performs a complete inference and description of the individual's trajectory.

In the trajectory description, TSI-LLM emphasizes the category to which the individual belongs before reasoning (the bolded font). Then, it describes the activities the individual engages in at specific times and locations in detail and explain the reason for each change in position. Finally, it summarizes the activity patterns of this occupational category (the red font). This result reflects the progressive relationship of the three tasks when TSI-LLM reasons about each occupational category. From the sequence of activities, it is evident that TSI-LLM's reasoning about individual activities changes with the occupational category, and it makes reasonable explanations in the trajectory description. It indicates that using the CoT strategy in the prompt allows TSI-LLM to perform step-by-step inference, enabling the model to capture the relationships between tasks, thus leading to logical results.

%% file: section/7_conclusion.tex
\section{Conclusion}
In this study, we develop TSI-LLM, a trajectory semantic inference framework based on Large Language Models. In our framework, the individual trajectory is constructed as spatio-temporal trajectory chain containing mobility information through spatio-temporal attributes enhanced data formatting. In addition, we design the trajectory semantic inference prompt, adopting the Chain-of-Thought strategy to enable LLMs to perform reasonable reasoning and logical explanation of the individual's occupational category, activity sequence, and trajectory description. The semantic inference experiments are carried out on the Shenzhen individual trajectory dataset, and the results proved the rationality and interpretability of the proposed method. However, there are still some limitations. A more quantitative analysis should be further conducted on large-scale trajectories semantic reasoning. We expect that this study is not just a technical work for trajectory semantic inference but also a way to generate a text-trajectory multimodal dataset for further research in human mobility analysis.

%% file: section/8_acknowledgement.tex
\section{Acknowlegdement}
This work was supported in part by the National Key R\&D Program of China (No.2021YFC2600505), Guangdong Basic and Applied Basic Research Foundation (No. 2024A1515012020) and National Natural Science Foundation of China (No. 42271474).